\documentclass[letterpaper, 10 pt, conference]{ieeeconf}  

\IEEEoverridecommandlockouts                              

\usepackage{flushend}
\usepackage{graphicx} 
\usepackage{amsmath} 
\usepackage{amssymb}  
\usepackage{caption}

\newcommand{\mvec}[1]{\mathbf{#1}}

\title{\LARGE \bf
	Predicting Out-of-View Feature Points for Model-Based\\[0.5ex]  Camera Pose Estimation
}

\author{Oliver Moolan-Feroze, Andrew Calway$^{1}$
\thanks{$^{1}$Authors are with the Department of Computer Science, University of Bristol, Merchant Venturers Building, Bristol BS8 1UB, United Kingdom.
{\tt\small \{om0000,andrew.calway\}@bristol.ac.uk}}%
}
\begin{document}

\makeatletter
\let\@oldmaketitle\@maketitle
\renewcommand{\@maketitle}{\@oldmaketitle
  \begin{center}
  \vspace*{4ex}
  \includegraphics[width=0.9\linewidth]
    {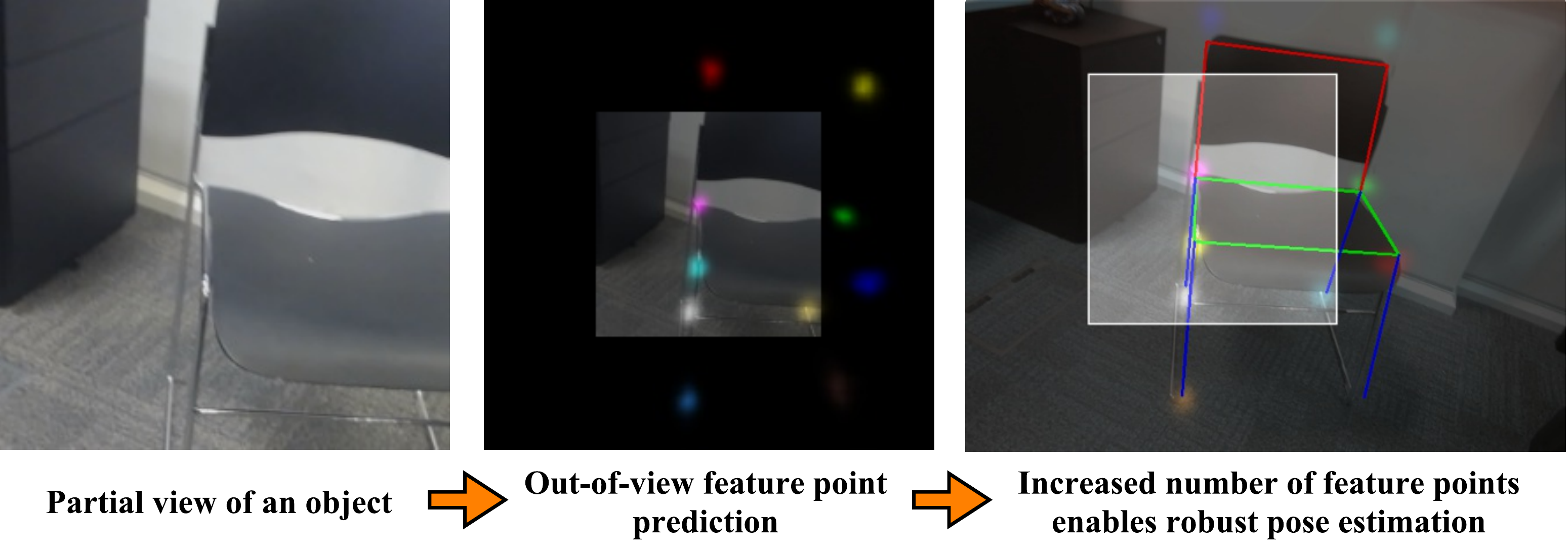}
	\captionof{figure}{Example showing camera pose estimation given an incomplete view of an object. Left) Incomplete view of a chair that has been cropped from a larger image. Centre) Chair feature points extracted using our CNN. Locations of features outside of the cropped area have been predicted. Right) Camera pose has been computed and the object model projected back into the original image.}
	\end{center}
	}%
\makeatother
\maketitle

\begin{abstract}
	In this work we present a novel framework that uses deep learning to predict object feature points that are out-of-view in the input image. This system was developed with the application of model-based tracking in mind, particularly in the case of autonomous inspection robots, where only partial views of the object are available. Out-of-view prediction is enabled by applying scaling to the feature point labels during network training. This is combined with a recurrent neural network architecture designed to provide the final prediction layers with rich feature information from across the spatial extent of the input image. To show the versatility of these out-of-view predictions, we describe how to integrate them in both a particle filter tracker and an optimisation based tracker. To evaluate our work we compared our framework with one that predicts only points inside the image. We show that as the amount of the object in view decreases, being able to predict outside the image bounds adds robustness to the final pose estimation.  
\end{abstract}

\section{INTRODUCTION}\label{sec:introduction}

\bstctlcite{IEEEexample:BSTcontrol}
Presented with an incomplete view of an object, a human is able to make predictions about the structure of the parts of the object that are not currently visible. This allows us to safely navigate around large objects and enables us to predict the effects of various manipulations on smaller objects. Both of these skills are important areas of investigation for robotics researchers. In this work, we look at the problem of out-of-view prediction in the context of model-based tracking, where the camera pose (location and orientation) is estimated from incomplete views of a known object. In a typical model-based tracker, features are extracted from an image in the form of points, lines, or other higher level cues. By matching these features to a representation of the tracked model, a estimate of the camera pose can be computed. Key to this process is that a sufficient number of features are extracted so as to be able to get a robust pose estimate. When only a partial view of the object is available, the number of visible features is reduced and consequently tracking performance is affected. By predicting out-of-view, we expand the set of possible correspondences, and increase the robustness of the tracking.

One area where this is particularly important is for autonomous inspection. In this application, it is often necessary for the inspection platform to be close to the surface to collect useful data. This results in large areas of the structure falling out-of-view of the inspection platform's cameras. Being able to predict this missing structure will enable a more robust tracking, which in-turn provides safe navigation as well as making possible post-inspection data processing such as image stitching and 3D reconstruction.

The main contribution of this work is our out-of-view feature point prediction method which is based on a Convolutional Neural Network (CNN). The architecture is modelled around the encoder-decoder design where the output of the network is a set of heatmaps where a higher intensity value corresponds to a higher confidence of feature point location. Typically these networks produce a direct relationship between locations on the input image and locations on the output heatmap. We proposed to break this relationship by scaling the labels to bring a greater extent of the object into the heatmaps than is contained within the input image.

In Section~\ref{sec:related-work} we review the previous literature on the use CNNs in both tracking and feature point extraction. In Section~\ref{sec:method} we detail our method for feature point prediction using CNNs and how these predictions can be integrated into a tracking system. In Section~\ref{sec:experiments} we present a set of experiments to show how out-of-view predictions produce more robust pose estimates when dealing with partial views and present some example results of the predictions integrated into a particle filter tracker. Finally, in Section~\ref{sec:conclusion} we give some conclusions and our goals for future work.

\section{RELATED WORK}\label{sec:related-work}

The research applying deep learning to camera pose estimation can be split into two groups: those that take an end-to-end approach where pose is regressed directly from an input image, and those that use deep learning as an intermediary step that can be integrated into a traditional tracking framework. Of the end-to-end group, the first work to tackle this is in~\cite{kendall_posenet:_2015}. The authors build a end-to-end pose regression network named PoseNet which consists of a convolutional part, based in the VGG~\cite{simonyan_very_2014} style, with two dense blocks appended to regress the translation and rotation. This is trained using views accompanied with pose labels computed using structure from motion (SFM). Through leveraging transfer learning, they are able to train a network which can predict pose in both indoor and outdoor scenes using only a small number of labelled images. The authors expand on this in~\cite{kendall_modelling_2015}, where they explore the use of Bayesian Deep Learning to provide a measure of uncertainty alongside the pose estimate. Having uncertainties is beneficial in a localisation framework as they provide a means to reject estimates as incorrect if the certainty is low. In~\cite{kendall_geometric_2017}, this work is further expanded through the use of a novel loss function. In the previous PoseNet methods, the network was trained by minimising the mean squared error between the network output and the label, where the label contains an translation and a rotation represented as a quaternion. As the difference in scale of these two values can be large, an arbitrary weighting factor is used when combining the differences in the loss function. To address this, the authors train the network in a ``geometrical'' way by minimising the mean squared differences between scene points when projected through the network output and label using a pinhole camera model. This removes the need for arbitrarily chosen weighting values.  

The end-to-end pose regression has been adapted by Clark et.al.~\cite{clark_vidloc:_2017} to take advantage of the temporal smoothness between video frames to improve pose estimates. The authors stack a series of Long Short-term memory (LSTM) layers after the convolutional part of the network, which are able to integrate features from previous time steps to improve the robustness of the pose estimates. They show that this information provides large accuracy improvement over the original PoseNet method. Recurrent neural networks (RNNs) are also used in~\cite{wang_deepvo:_2017} where they are applied to visual odometry. The authors state that the recurrent units implicitly learn the motion dynamics of the system, eliminating the need for complex geometrical pipelines. 

At this time, there remain a number of problems with the end-to-end learning of camera pose. The most prohibitive of these is that pose-labelled images are need to train the network. This data is often hard to obtain, and a sufficient amount of it is needed to train models that will generalise well. Furthermore, as shown in the evaluation of the cited papers, traditional geometric based tracking methods still outperform the end-to-end models.

The work in~\cite{Pavlakos} addresses these problems by choosing not to directly regress the pose of the camera, and instead uses a CNN to extract model feature points in the form of a set of heatmaps. The peaks of the heatmaps are chosen as feature locations, and the values of the peaks indicate the location uncertainty. Given a known 3D model, the pose of the camera is then estimated through a minimization process. To generate the heatmaps the authors borrow the stacked hourglass network architecture~\cite{newell_stacked_2016}, which combines multiple encoder-decoder networks one after the other. This enables the learning of long range relationships between feature points. This network architecture was initially proposed for the task of human pose estimation which is where we find the state-of-the-art in feature point estimation.  

The use of CNNs in the extraction of joint locations for human pose estimation is well established. Except for the work in~\cite{toshev_deeppose:_2014}, which directly regresses the $\left(x,y\right)$ locations of the joints, the majority of the methods -- as well as our own work --  produce heatmaps of the locations. As explained in~\cite{sun_compositional_2017}, the reason for this is that during training the direct regression method does not internalise as well the spatial relationships between the points. Indeed, much of the literature in this field is aimed at producing methods that can leverage the spatial relationships to produce accurate predictions. This is achieved in~\cite{newell_stacked_2016} by the stacking of multiple encoder-decoder networks. In~\cite{wei_convolutional_2016}, the authors propose a sequential architecture that applies multiple CNNs one after the other to iteratively improve the point predictions. This is combined with intermediate supervision of the learning at each stage to improve training. Carreira et. al.~\cite{carreira_human_2015} address this through the use of an iterative process which fine-tunes the prediction output over a number of iterations. Each iteration seeks only to make small positive corrections to the output of the previous iteration. In~\cite{pfister_flowing_2015}, another iterative method is proposed which uses temporal information from video in the form of optical flow fields to improve point prediction over a number of frames. One of the problems with a stacked or iterative method is the extra computational overhead required, which depending on the size of the network can be prohibitive for a tracking based system.

\section{METHODS}\label{sec:method}

\subsection{Method Overview}

In this section we will describe our method for camera pose estimation from partial views of an object. Similar to the work in~\cite{Pavlakos}, our method is split into two parts. First, given an image containing an incomplete view of the tracked object, 2D feature points corresponding to 3D locations on the object are extracted from the image using a CNN. Second, the locations of the 2D features are used to compute the pose of the camera. As the focus of this work is predicting out-of-view feature points, the majority of this section is devoted to the first part. We do however describe two methods by which the predictions can integrated into a tracking framework: one based on a particle filter, and one based on direct optimisation. 

\subsection{Predicting out-of-view feature points using CNNs}\label{sec:method-predicting}

In our system, we represent the tracked object as a set of 3D points $\mvec{P}^{m} = \left\{ p^{m}_1, \ldots, p^{m}_n \right\} \in \mathbb{R}^3$. These points are chosen so as to correspond to easily identifiable locations as well as to provide enough 3D structure to be able to robustly estimate the pose. Using a pinhole camera model with a translation $\mathbf{t}$, rotation $\mathbf{R}$, and camera intrinsics $\mathbf{K}$ we can project the model points $\mathbf{P}^m$ to a set of corresponding 2D points $\mathbf{P}^c = \left\{p^c_1, \ldots, p^c_n \right\} \in \mathbb{R}^2$  on the image plane using
\begin{equation}\label{eq:projection}
	p^{c}_i =  \mathbf{K}\left[\mathbf{R} | \mathbf{t}\right] p^m_i \enspace .
\end{equation}
In a model-based tracking system, the goal is to extract the locations of $\mathbf{P}^c$ and from these, compute the translation $\mathbf{t}$ and rotation $\mathbf{R}$. 

\begin{figure}[t]
	\centering
	\includegraphics[width=\linewidth]{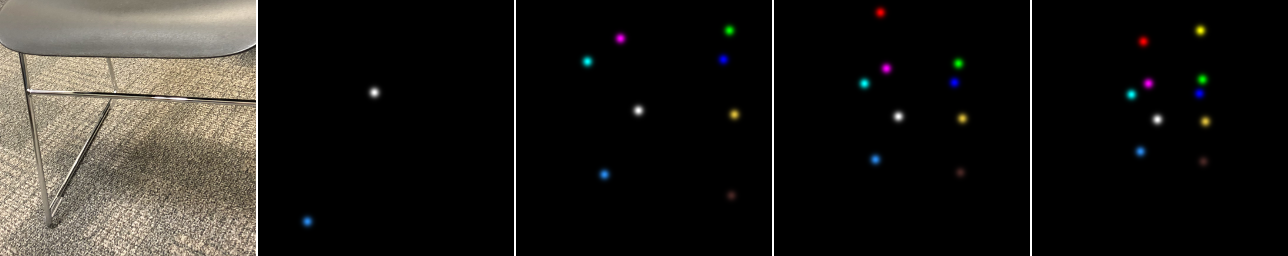}
	\caption{Example labels produced by different $s$ values. On the left is the input image showing a partial view of a chair. To the right are labels that are produced using $s$ values of $1$, $\frac{1}{2}$, $\frac{1}{3}$, $\frac{1}{4}$. The smaller the scale value, the larger number of out-of-view points are brought into the label.}
	\label{fig:scaled-labels}
\end{figure}

As stated in Section~\ref{sec:introduction}, the aim of this work is to enable the prediction of points that lie out-of-view of the input image. To this end, we use a CNN which takes an  RGB image $I$ as input and produces a set of heatmaps $\mathbf{H} = \left\{ h_1, \dots, h_n \right\}$, where each map $h_i$ corresponds to a different feature point $p^c_i$. The type of heatmap produced by the CNN in~\cite{Pavlakos} have a direct spatial relationship to the input image. That is, the 2D pixel coordinates of a feature on the image should be the same as the coordinates of the peak in the heatmap. This is achieved during the training process. For each training image, a heatmap is produced by placing a 2D Gaussian on the image coordinates of each feature. In our work, to force the network to predict out-of-view points, we produce the heatmaps differently by generating a new set of feature locations $\mathbf{P}^s = \left\{ p^s_i, \ldots, p^s_n \right\}$ by applying a scaling and offset determined by a value $s$
\begin{equation}\label{eq:projection-scaled}
	p^s_i = s \cdot p^c_i + \frac{\left(N - s \cdot N\right)}{2} \enspace ,
\end{equation}
where $N$ is the dimensions of the input image. These new points are then taken as the 2D locations of the Gaussians used to create the heatmaps. The effect of this operation is to reduce the size of the object within the heatmap, which consequently bring more of objects points into view. The value $s$ can be seen as a zooming operation, where values of $s$ less than 1 will produce heatmaps that contain more of the object than the input image. Given $s=1$, labels will be produced that are the same as the work in~\cite{Pavlakos} and will retain the one-to-one spatial relationships. Examples of different labels corresponding to different values of $s$ can be seen in Fig.~\ref{fig:scaled-labels}.

The type of CNN architecture that we use is based on the encoder-decoder style. The encoder part consists of a series of convolutional filters followed by max-pooling layers which sequentially reduce the resolution of the input and draw in feature information from an increasingly greater spatial area of the image. The deconvolutional part of the network takes the filter activations at the smallest resolution and applies a set of linear upsampling layers followed by more convolutions which increase the spatial resolution back to its original size. This type of network is commonly used in semantic segmentation methods such as~\cite{Badrinarayanan} as well as joint location prediction for human pose estimation~\cite{newell_stacked_2016}. 

One of the key effects of this type of network is the accumulation of features from a broad area of the input image during the encoding section, which can go on to influence the predictors in the final layer of the decoder. In our application where only partial views of the object are visible, this is doubly important, as often, only a small amount of useful visual information is contained within the image. To be able to make robust predictions it is important that the decoder section of the network has access to as much of this information as possible. In a stacked network or an iterative network, the long range information is incorporated by producing and operating on multiple intermediate predictions. However, as more networks are stacked, or more iterations performed, the amount of time it takes to make predictions increases, making real time tracking impossible. Another method of increasing the amount of feature information available is to have the encoder greatly reduce the spatial resolution of the input before applying the decoding. However, we found that this reduces the robustness of the predictions as well as increasing processing time.  
\begin{figure*}[t]
	\centering
	\includegraphics{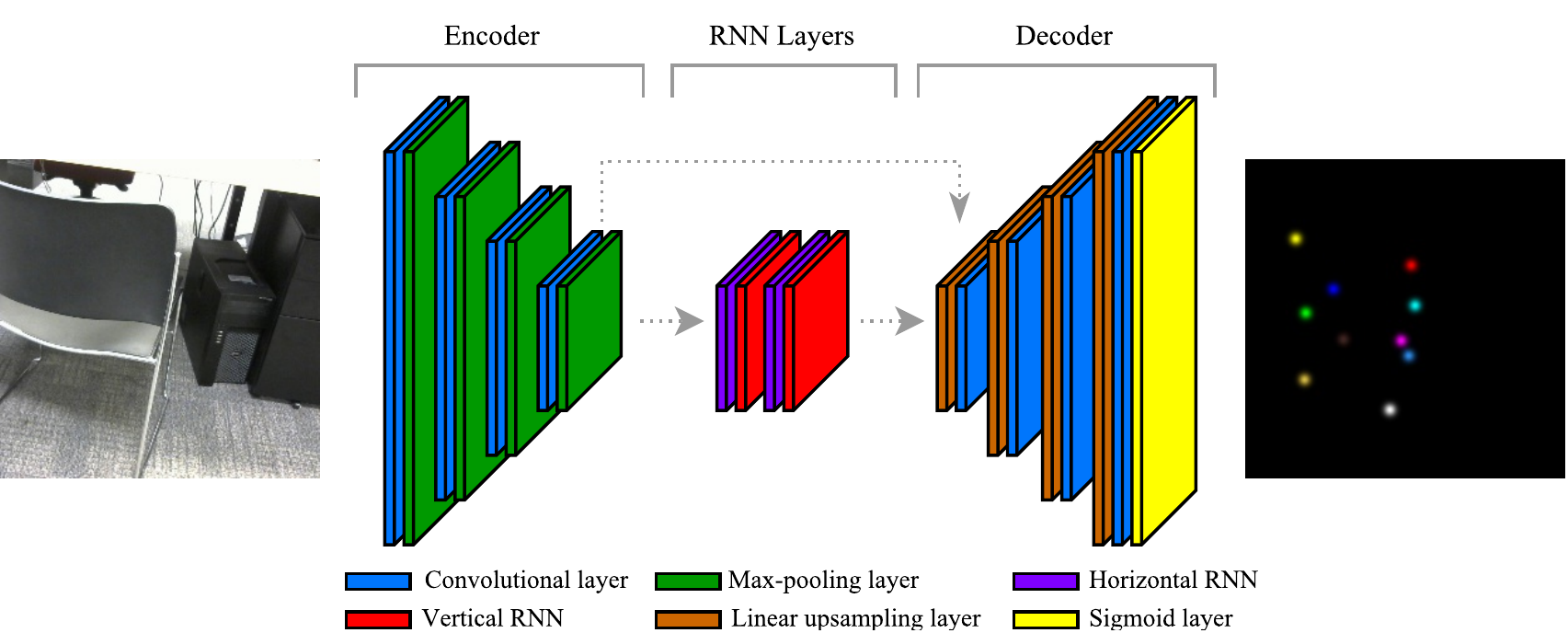}
	\caption{Diagram showing the architecture of the network. Input image is on the left and an RGB representation of the multi-channel output is on the right.}
	\label{fig:network-diagram}
\end{figure*}

To address this issue in a computationally inexpensive way we took inspirations from the work in~\cite{Visin2016} and add a set of RNN layers to the network at the end of the encoder section and before the decoder. These RNNs named ReNet layers comprise a series of horizontal RNNs which sweep the features left and right, followed by vertical RNNs which sweep the features up and down. During training, these layers learn which features are important and can pass them onto difference spatial locations as needed. This provides the decoder part access to features from any area of the input image without adding excessive computational overhead or adversely affecting prediction performance. We make one change to the implementation of the ReNet layers as presented in~\cite{Visin2016} in that we concatenate the output of the encoder onto the output of the ReNet layers. We found that without doing so, the network was unable to train. A diagram giving an overview of the network is shown in Fig.~\ref{fig:network-diagram}. The details of the network can be found in Table~\ref{tab:cnn-details}

\begin{table}[t]
	\caption{CNN network details}\label{tab:cnn-details}
	\centering
	\begin{tabular}{ | c | }
		\hline
		Conv 3x3 (64), Batchnorm, ReLU    \\ 
		Max-pooling 2D \\
		\hline
		Conv 3x3 (128), Batchnorm, ReLU    \\ 
		Max-pooling 2D \\
		\hline
		Conv 3x3 (256), Batchnorm, ReLU    \\ 
		Max-pooling 2D \\
		\hline
		Conv 3x3 (512), Batchnorm, ReLU    \\ 
		Max-pooling 2D \\
		\hline
		Horizontal Bi-directional GRU, (256) \\
		Vertical Bi-directional GRU, (256) \\
		\hline
		Horizontal Bi-directional GRU, (256) \\
		Vertical Bi-directional GRU, (256) \\
		\hline
		Upsampling 2D \\
		Conv 3x3 (512), Batchnorm, ReLU    \\ 
		\hline
		Upsampling 2D \\
		Conv 3x3 (256), Batchnorm, ReLU    \\ 
		\hline
		Upsampling 2D \\
		Conv 3x3 (128), Batchnorm, ReLU    \\ 
		\hline
		Upsampling 2D \\
		Conv 3x3 (64), Batchnorm, ReLU    \\ 
		\hline
		Conv 1x1 (num classes) \\ 
		Sigmoid \\
		\hline
	\end{tabular}
\end{table}

\subsection{Tracking}\label{sec:method-tracking}

In this section we will describe how we use the feature point predictions with two different tracking systems. The simplest way of doing this is through the use of a particle filter. In this system, at time step $t$ the pose of a camera is defined by a set of particles, with each particle providing a 3D translation $\mathbf{t}$ and 3D rotation $\mathbf{R}$, represented as a quaternion $\mathbf{q}$. The pose is accompanied by a support value $w$, which indicates the degree of confidence in the pose. A singular pose estimate can be obtained through a weighted average over the particles. 

At each time step, the particles are updated through the use of a motion model, and for each updated particle the weight is recomputed. To evaluate the weight of the particle we use the heatmaps produced by the CNN. We apply (\ref{eq:projection}) to the model points $\mathbf{P}^m$ to get the 2D points $\mathbf{P}^c$ and then apply (\ref{eq:projection-scaled}) using the $s$ value we chose during training to get $\mathbf{P}^s$. We then find the heatmap values at the coordinates of the points in $\mathbf{P}^s$ and sum them
\begin{equation}
	w = \sum_{i}^{n} h_i\left(p^s_i\right) \enspace .
\end{equation} 
If the projected points lie on a location that has been predicted to be the location of that feature, it will add a high value to the weight. Conversely, if a point is projected to a location where the network hasn't predicted a feature, a small value will be added to the weight. After computing the weights, importance sampling is used which removes particles that have with low confidence and replicates particles with high confidence. The benefit of using a particle filter is that it is simple to implement, fast, and integrates well with the heatmap type of feature point prediction.

The second tracking method that we present is based on an optimiser. For each image, we compute the predictions using the CNN. Each heatmap is then normalised such that they sum to one and then we apply the negative log. This converts the heatmaps into cost images where the smaller the value, the more likely it corresponds to a location of a particular feature point. We then compute the pose as a minimisation problem
\begin{equation}
	\left[\mathbf{R},\mathbf{t}\right]^* = \arg\min_{\left[\mathbf{R}, \mathbf{t}\right]} \Phi\left(\mathbf{R}, \mathbf{t}, \mathbf{P}^m, \mathbf{K}\right) \enspace,
\end{equation}
where $\Phi$ is the cost function, which is evaluated by projecting the points onto the negative log cost images, and summing the values at the location of the points. The function will be at a minimum when the projected points align with the predictions from the CNN. We optimise the function using gradient descent. This method is beneficial to the particle filter in that it provides a single pose estimate. We did find however that it requires a decent initialisation of $\mathbf{t}$ and $\mathbf{R}$ and it is necessary to smooth the prediction images to increases the long range support during optimisation.

\section{EXPERIMENTS}\label{sec:experiments}

To evaluate the performance of our work, we conducted two sets of experiments. The first is outlined in Section~\ref{sec:experiments-pose-recovery} and aims to empirically evaluate the ability of our method to estimate the pose of the camera given views of an object with varying levels of incompleteness. We compare a number of versions of the proposed network each with difference scale values $s$ as well as the network proposed in~\cite{Pavlakos}. The second set of experiments detailed in Section~\ref{sec:experiments-tracking} examines the tracking performance when the out-of-view predictions are integrated with a particle filter framework. As we do not have ground truth poses for the tracking sequences we show the results visually through reprojections of the object model back onto the input image.


For our experiments, we used 3 different objects. For the first experiment we use a chair and for the second we add a computer monitor, and a computer keyboard. The chair has 10 feature points, the screen and keyboard each have 4, one for each corner. The reason for including both the screen and keyboard was to evaluate the method when the object has only a small number of feature points. To train the networks we capture images of each object from a number of different views using a standard USB webcam. For the chair we obtained $\sim250$ images and for the screen and keyboard we captured $\sim100$ images. The images were split $80-20\%$ to provide a test set so we know when to stop training. During training we augment the images using random translations, rotations and scaling. The networks were optimised using the Adam optimiser~\cite{kingma_adam:_2014} with a learning rate 0f $0.05$. As suggested in~\cite{Badrinarayanan}, a dropout of $0.2$ was used in the final 2 layers of the encoder and first 2 layers of the decoder to reduce model overfitting. The networks were trained using the PyTorch framework on a Nvidia GeForce GTX 1060 with training lasting until test set loss plateaued. This took around 3 days for the chair network and around 2 days for the screen and keyboard.

\begin{figure*}[t]
	\centering
	\includegraphics[width=\linewidth]{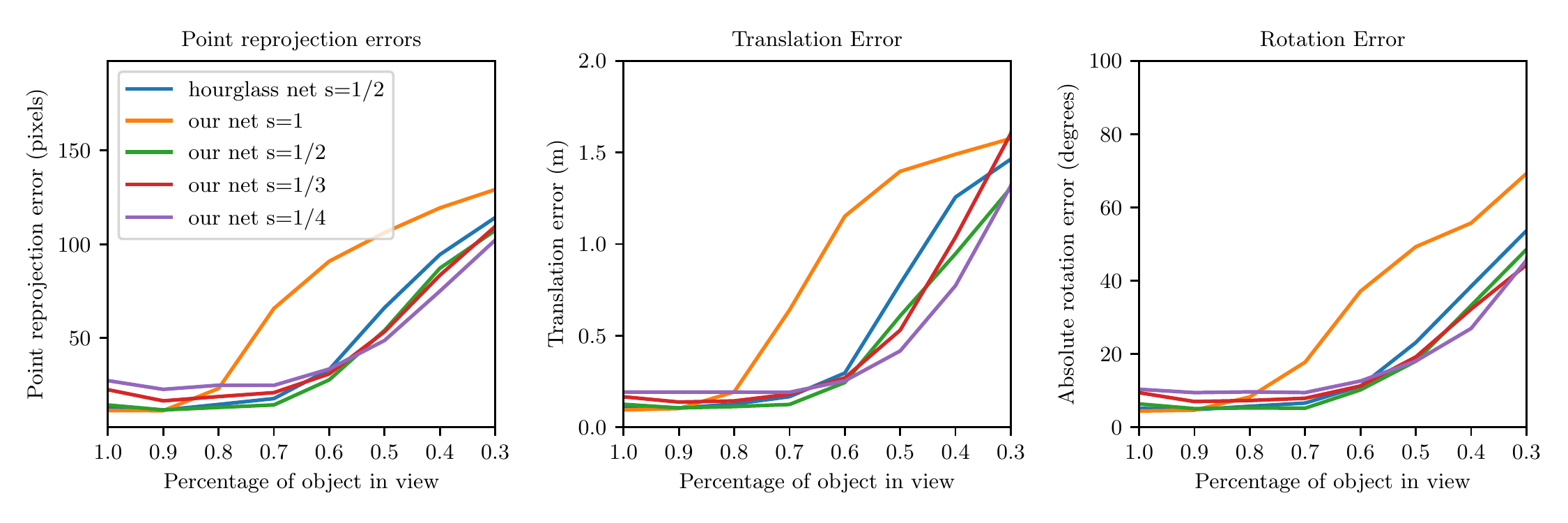}
	\caption{Plots of the performance of the different network types for reprojection error, translation error and rotation error. For each view, the amount of object visible was rounded to the nearest 0.1 percent and the median error was computed from each group.}
	\label{fig:pose-eval-test}
\end{figure*}

\subsection{Pose Recovery Evaluation}\label{sec:experiments-pose-recovery}

\begin{figure*}[t]
	\centering
	\includegraphics[width=\linewidth]{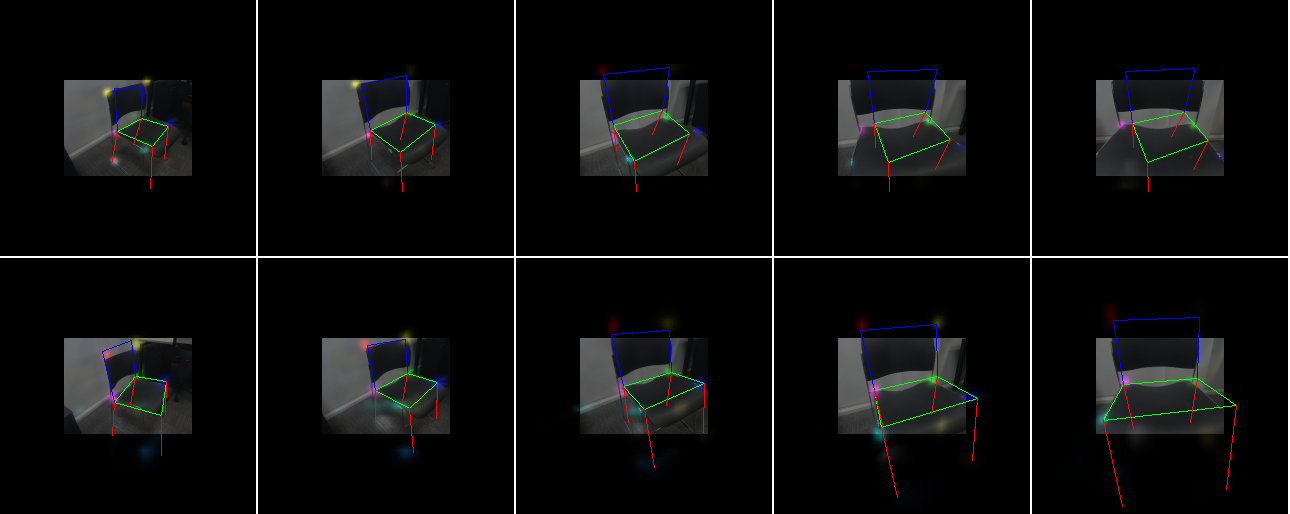}
	\caption{Examples of the tracking results. The estimated pose is used to project the model into the image. Top) Tracking results where the scale value $s=1$. This means that the network is not trying to predict outside of the image. Bottom) Tracking on the same sequence where scale value $s=2$}
	\label{fig:tracking-results}
\end{figure*}

To evaluate the performance of the system to recover pose given incomplete views we acquired a set of images using a USB webcam (a different camera than the one used to capture the training data). Each of the views contained the entirety of the object and were taken from a number of different viewpoints. For each view we then manually landmarked the feature points. Using these and the camera intrinsics we applied the EPnP algorithm~\cite{lepetit_epnp_2008} to compute the camera pose for each image. As we wanted to evaluate the performance using images with varying amounts of the object in view, we applied random transforms to image. These transforms included rotations, translations and scaling. Finally, we cropped the transformed images to a size of $256$x$256$. To determine the amount of the object contained within the image, we computed the convex hull of the object feature points after the transformation. We then extracted the area of the hull that was within the image bounds and divided it by the total area of the hull. This gave us a percentage value which ranged from $1.0-0.3$ for each view. In total we extracted 2500 views using this method. To estimate the pose of the camera from each view, we first apply the network and compute the feature point predictions. Next, the prediction image is transformed using the inverse of the transform applied to create the view. The reason for this is that the transformation process applied to create a view make the camera intrinsics incorrect. After applying the inverse transform to the predictions, we compute the pose  through the optimisation process detailed in Section~\ref{sec:method-tracking}. The pose estimates were then compared to the poses computed from the manual landmarks using the EPnP algorithm. In this experiment we evaluated 4 versions of our own network with $s = 1, \frac{1}{2}, \frac{1}{3}, \frac{1}{4}$. We also compared our work with the network proposed in~\cite{Pavlakos}, which we trained on the same data and set the label scaling to $\frac{1}{2}$.

\begin{figure*}[t]
	\centering
	\includegraphics[width=\linewidth]{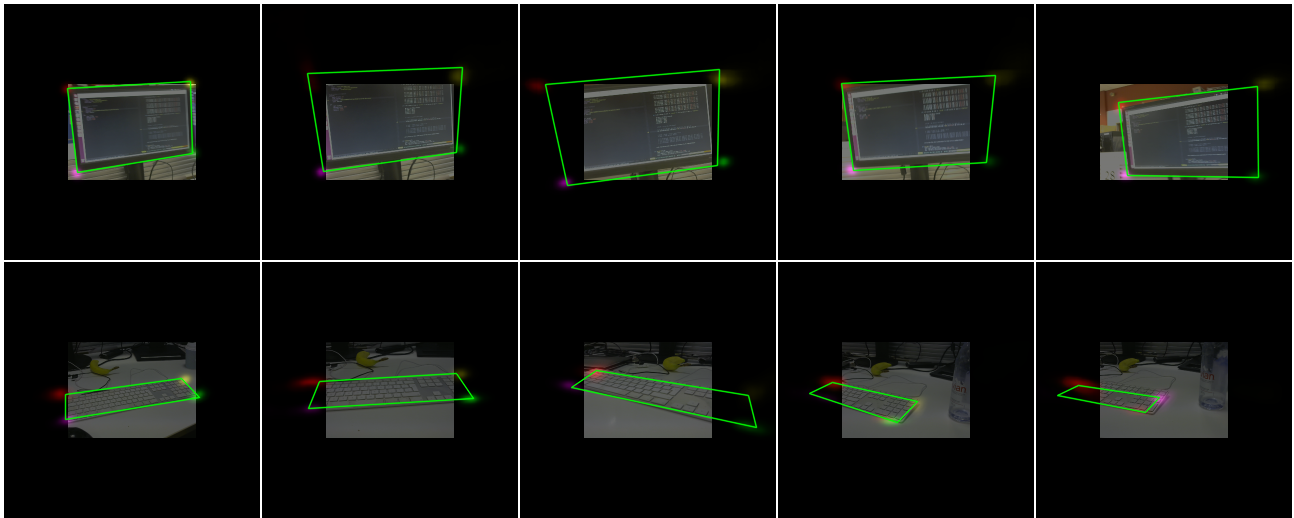}
	\caption{Tracking results for the other objects. Top) Computer screen and Bottom) Computer keyboard}
	\label{fig:tracking-results-ks}
\end{figure*}

To quantify the pose accuracy for the different networks we directly compare the ground truth with the estimates and compute the error in the translation as well as the absolute rotation error. We also compute the difference between the projection of the model points using the ground truth and estimated pose values. The results of these evaluations can be seen in Fig.~\ref{fig:pose-eval-test}. 

The results show that for all the methods evaluated, the network trained with $s=1$ is the least robust to reductions in the visible percentage of the object, with performances dropping sharply with percentages lower than 0.8. The comparison of this to the networks trained with $s < 1$ -- where performance drops off slower and at a lower percentage -- shows the usefulness of predicting out-of-view feature points when dealing with incomplete views. When comparing the performances of the methods trained with $s < 1$, the conclusions are less clear. We can see that the performance of the network proposed in~\cite{Pavlakos} when trained with $s=\frac{1}{2}$ is slightly worse than the performance of our network with the same $s$ value, especially at the lower percentages. This shows that our architecture is better suited to this particular problem. In addition, our network is considerably faster than the stacked hourglass architecture. When looking at the networks trained with the smallest values of $s$ we can see that they are slightly more robust at the lowest percentages; this makes sense as they are predicting the largest range of out-of-view features. However, the smaller $s$ values perform slightly worse at higher percentages, which we believe is due to the reduced resolution of the heatmaps that are a result of the scaling process. These results show that there is a trade-off when choosing an $s$ value, which depends on the expected percentage of the object that is likely to be viewed by the system.


\subsection{Tracking Examples}\label{sec:experiments-tracking}

In this experiment we aim to show the effect of out-of-view feature point prediction in a tracking framework. To do this we used two versions of the chair model, one trained with $s=1$ and one trained with $s=\frac{1}{2}$. We also trained networks for the screen and keyboard, each with $s=\frac{1}{2}$. To perform the tracking we implemented the particle filter as described in Section~\ref{sec:method-tracking}. To show the performance, we applied the tracker to a set of video sequences of the different objects. At a number of different time steps we used the estimated camera pose to project the models into the original image. The tracking was done on a laptop, with the particle filter running on an Intel i7 processor, and the network running on a Nvidia GeForce GTX 1050. For all examples the tracker ran in real-time with $\sim$30 fps.

Examples from the sequences for the chair can be seen in Fig.~\ref{fig:tracking-results} and for the Keyboard and Screen in Fig.~\ref{fig:tracking-results-ks}. For the chair example, the top row of images are produced with $s=1$ and the bottom with $s=\frac{1}{2}$. We can see that at the start, where the majority of the chair is in view, both methods are able to calculate a good pose. However, as the camera moves closer to the chair, the performance of the $s=1$ model deteriorates and tracking is lost. For the model where $s=\frac{1}{2}$, tracking remains good. For the screen and keyboard objects, the tracking is arguably harder, as there are less feature points to use for tracking. However, the images show that even in instances where all the corner points fall outside of the image, tracking is still possible.

\section{CONCLUSIONS}\label{sec:conclusion}

In this work, we have presented a novel method for predicting out-of-view feature points using CNNs with the aim of enabling camera pose estimation given incomplete views of an object. We present a tailored CNN architecture that is able to integrate rich feature information from across the input image, allowing feature point prediction from this challenging type of data. In our evaluation we show that the ability to predict feature points outside of the input image adds robustness to the pose computation as the amount of visible object is reduced. We have also shown that in a tracking scenario, out-of-view point prediction enables tracking to continue when in-view only prediction does not.  

For future work we are interested in expanding our research to deal with class based object tracking. That is, being able to predict object features for all types of an object rather than just one particular instance. In addition, we aim to investigate the more challenging problem of feature point prediction for articulated and deformable objects.

\addtolength{\textheight}{-13cm}  

\bibliographystyle{IEEEtran}
\bibliography{IEEEabrv,root} 

\end{document}